\documentclass[twoside,11pt]{article}
\usepackage{jair, theapa, rawfonts}
\usepackage{textcomp} %luo
\usepackage[T1]{fontenc} %luo, for underscore symbol
\usepackage{amssymb} %luo
\usepackage{amsmath,bm} %luo
\usepackage{epsfig} %luo
\usepackage{flushend} %luo
\usepackage{ntheorem} %luo
\usepackage{times}
\usepackage{helvet}
\usepackage{courier}
\usepackage[linesnumbered,ruled,vlined]{algorithm2e}%%shaowei, for algorithm
\usepackage{longtable}
\usepackage{multirow}
\usepackage{supertabular}
\usepackage{multicol}
\usepackage{courier}
\usepackage{color}
\usepackage{array}

\usepackage{graphicx}
\usepackage{epstopdf}

\newcommand{\PreserveBackslash}[1]{\let\temp=\\#1\let\\=\temp}
\newcolumntype{C}[1]{>{\PreserveBackslash\centering}p{#1}}
\newcolumntype{R}[1]{>{\PreserveBackslash\raggedleft}p{#1}}
\newcolumntype{L}[1]{>{\PreserveBackslash\raggedright}p{#1}}
\newtheorem{definition}{\bf Definition}
\newtheorem{proposition}{\bf Proposition}

\ShortHeadings{Local Search for Minimum Weight Dominating Set}
{Wang, Cai, \& Yin}
\firstpageno{267}

\begin{document}

\title{Local Search for Minimum Weight Dominating Set with Two-Level Configuration Checking and Frequency Based Scoring Function}

\author{\name Yiyuan Wang \email yiyuanwangjlu@126.com \\
       \addr School of Computer Science and Information Technology\\
        Northeast Normal University, Changchun, China
       \AND
       \name Shaowei Cai \email caisw@ios.ac.cn \\
       \addr State Key Laboratory of Computer Science, Institute of Software \\
       Chinese Academy of Sciences, Beijing, China
       \AND
       \name Minghao Yin* \email ymh@nenu.edu.cn \\
       \addr School of Computer Science and Information Technology\\
       Northeast Normal University, Changchun, China\\
       corresponding author}

% For research notes, remove the comment character in the line below.
% \researchnote

\maketitle

\begin{abstract}
The Minimum Weight Dominating Set (MWDS) problem is an important generalization of the Minimum Dominating Set (MDS) problem with extensive applications.
This paper proposes a new local search algorithm for the MWDS problem, which is based on two new ideas. The first idea is a heuristic called two-level configuration checking (CC$^{2}$), which is a new variant of a recent powerful configuration checking strategy (CC) for effectively avoiding the recent search paths. The second idea is a novel scoring function based on the frequency of being uncovered of vertices. Our algorithm is called CC$^{2}$FS, according to the names of the two ideas.
The experimental results show that, CC$^{2}$FS performs much better than some state-of-the-art algorithms in terms of solution quality on a broad range of MWDS benchmarks.
\end{abstract}

\section{Introduction}
\label{Introduction}
Given an undirected graph $G$, a dominating set $D$ is a subset of vertices such that every vertex not in $D$ is adjacent to at least one member of $D$. 
The Minimum Dominating Set (MDS) problem consists in identifying the smallest dominating set in a graph. The Minimum Weight Dominating Set (MWDS) problem is a generalized version of MDS. In the MWDS problem, each vertex is associated with a positive value as its weight, and the task is to find a dominating set that minimizes the total weight of the vertices in it.

%\citeyear{ref66}

The MWDS problem has played a prominent role in various real-world domains such as social networks, communication networks, and industrial applications. For example, Houmaidi et al. make the first known attempt to solve the sparse wavelength converters placement problem, where the MWDS problem is used to reduce the number of full wavelength converters in the deployment of wavelength division multiplexing all-optical networks \cite{ref70}. In the work of Subhadrabandhu, Sarkar, and Anjum \citeyear{ref66}, the MWDS problem is used in determining the nodes in an adhoc network where the intrusion detection software for squandering detection needs to be installed. Shen et al. propose a new method for multi-document by encoding this problem to the MWDS problem \cite{ref4}.
Also, the problem of gateways placement, which places a minimum number of gateways such that quality-of-service requirements are satisfied, can be encoded into the MWDS problem and effectively solved by MWDS algorithms \cite{ref3}. An important problem in Web databases is to find an optimal query selection plan, which has been proved to be equivalent to finding a MWDS of the corresponding database graph \cite{ref67}.
%The MWDS problem has played a prominent role in various real-world domains such as social network, communication network, and industrial applications \cite{ref3,ref67,ref4}.%ref3,ref70,ref66,

The problems MDS and MWDS have been proved to be NP-hard \cite{ref1,ref5}, which means that, unless P=NP, there is no polynomial-time algorithm for these problems. Several approximation algorithms have been introduced to solve the MWDS problem. An early constant-factor approximation algorithm for the MWDS problem in unit disk graphs is presented \cite{ref44}. A new centralized and distributed approximation algorithm for the MWDS problem is proposed with application to form a backbone for adhoc wireless networks \cite{ref76}. Dai and Yu propose a (5+$\varepsilon$)-approximation algorithm to form a MWDS for UDG \cite{ref43}. A (4+$\varepsilon$)-approximation dynamic programming algorithm for the MWDS problem is offered, which is the best approximation factor for unit disk graph without smooth weights \cite{ref15}. The first polynomial time approximation scheme for MWDS with smooth weights on unit disk graphs is introduced \cite{ref23}, which achieves a (1+$\varepsilon$)-approximation ratio, for any $\varepsilon$>0.

Nevertheless, as is usual, approximation algorithms with guaranteed approximation ratios do not have good performance in practice.
Most practical algorithms for solving the MWDS problem are heuristic algorithms. In the work of Jovanovic, Tuba, and Simian \citeyear{ref16}, an ant colony optimization (ACO) algorithm for MWDS is proposed, which takes into account the weights of vertices being covered. An algorithm called ACO-PP-LS uses an ant colony optimization method by considering the pheromone deposit on the node and a preprocessing step immediately after pheromone initialization \cite{ref17}. A hybrid steady-state genetic algorithm HGA is proposed by using binary tournament selection method, fitness-based crossover and a simple bit-flip mutation scheme \cite{ref17}. In the work of Nitash and Singh \citeyear{ref18}, a swarm intelligence algorithm called ABC uses an artificial bee colony method to solve MWDS. A hybrid approach EA/G-IR is presented, which combines an evolutionary algorithm with guided mutation and an improvement operator \cite{ref75}. In the work of Bouamama and Blum \citeyear{ref90}, a randomized population-based iterated greedy approach R-PBIG is proposed to tackle MWDS, which maintains a population of solutions and applies the basic steps of an iterated greedy algorithm to each member of the population.
However, the efficiency of existing heuristic algorithm are still not satisfactory, especially for hard and large-scaled instances (as will be shown in our experiments). The reason may be that the heuristic functions used in previous algorithms do not have enough information during the search procedure, and the cycling search problems can not be overcome by most algorithms as well.

%They return poor-quality solutions for hard instances and large-scaled instances, and even fail to return a solution for massive graphs (as will be shown in our experiments).

In this paper, we develop a novel local search algorithm for the MWDS problem based on two new ideas. The first idea is a new variant of the Configuration Checking (CC) strategy.
Initially proposed in the work of Cai, Su, and Sattar \citeyear{ref37}, the CC strategy aims to reduce the cycling phenomenon (i.e., revisiting candidate solutions which have been recently visited) in local search, by considering the circumstance of the solution components, which is formally defined as the {\it configuration}.
The CC strategy has been successfully applied to a number of well-known combinatorial optimization problems, including Vertex Cover \cite{ref37,ref27,ref49,ref93}, Set Covering \cite{ref83,ref89}, Clique problem \cite{ref87}, Boolean Satisfiability \cite{ref39,ref40,ref85,ref29} %,ref56,ref84
and Maximum Satisfiability \cite{ref81,ref82}, as well as application problems such as Golomb Rulers optimization \cite{ref86}.
%The CC strategy has been successfully applied to a number of well-known combinatorial optimization problems, including minimum vertex cover \cite{ref37,ref27}, minimum set cover \cite{ref83}, maximum clique  \cite{ref87}, SAT \cite{ref40,ref29,ref84,ref85} %ref29, and MaxSAT \cite{ref81,ref82},
It is straightforward to devise the CC strategy for MWDS, which works as follows. For a vertex, it is forbidden to be added into the candidate solution if its configuration has not been changed after the last time it was removed from the candidate solution.
However, when applied to the MWDS problem, the CC strategy does not lead to good performance. The problem may be that the original CC strategy is too strict for solving MWDS, i.e. forbidding too many vertices to be selected, which limits the search area of the algorithm.
In this work, we propose a variant of the CC strategy based on a new definition of configuration. In this strategy, the configuration of a vertex $v$ refers to its two-level neighborhood, which is the union of the neighborhood $N(v)$ and the neighborhood of each vertex in $N(v)$. This new strategy is thus called two-level configuration checking (abbreviated as CC$^{2}$).

The second idea is a frequency based scoring function for vertices, according to which the score of each vertex is calculated. Local search algorithms for the MWDS problem maintain a candidate solution, which is a set of vertices selected for dominating. Then the algorithms will use a scoring functions to decide which vertices will be selected to update the candidate solution, where the scores of vertices indicate the benefit (which may be positive or negative) produced by adding (or removing) a vertex to the candidate solution.
Four greedy algorithms for the MWDS problem are developed by using four different scoring functions \cite{ref17}. After that, some scoring functions have been proposed recently \cite{ref17,ref18,ref75,ref90}. 
These functions are mostly designed based on the information of the graph itself, for example vertex degree and vertex uncovered neighbour weight. 
An augmented cost function is designed, whose costs are either predetermined or evaluated during search \cite{ref91}. 
In this work, we also introduce a scoring function based on dynamic information of vertices, i.e., the frequency of being uncovered by the candidate solution. This scoring function exploits the information of the search process and that of the candidate solution. 
Moreover, different from the augmented cost function, our function does not have parameters and thus can be easily adapted to solve other optimization problems.

By incorporating these two ideas, we develop a local search algorithm for the MWDS problem termed CC$^{2}$FS (as its two main ideas are CC$^2$ and Frequency-based Score). We carry out experiments to compare CC$^{2}$FS with five state-of-the-art MWDS algorithms on benchmarks in the literatures including unit disk graphs and random generated instances, as well as two classical graphs benchmarks namely BHOSLIB \cite{ref53} and DIMACS \cite{ref77}, and a broad range of real world massive graphs with millions of vertices and dozens of millions of edges \cite{ref65}. Experimental results show that CC$^{2}$FS significantly outperforms previous algorithms and improves the best known solution quality for some difficult instances.

%The remainder of this paper is organized as follows: we give some necessary background knowledge in Section 2. Then, in Section 3 we review the previous configuration checking and then propose a new configuration checking strategy CC$^{2}$ in Section 4. Additionally, we shall introduce a frequency based scoring function and design a vertex selection method in Section 5 and Section 6. After that, a novel local search framework CC$^{2}$FS shall be proposed in section 7. In Section 8, experimental evaluations and the analyses of the experimental results will be shown. Conclusions and future directions are made in the last section.
%
In the next section, we give some necessary background knowledge. After that, we propose a new configuration checking strategy CC$^{2}$, a frequency based scoring function, and a vertex selection method. Then, we propose a novel local search algorithm CC$^2$FS, followed with experimental evaluations and analyses. Finally, we give concluding remarks.

\section{Preliminaries}
An undirected graph $G=(V,E)$ comprises a vertex set $V=\{v_{1},v_{2},\dots,v_{n}\}$ of $n$ vertices together with a set $E=\{e_{1},e_{2},\dots,e_{m}\}$ of $m$ edges, where each edge $e=\{v,u\}$ connects two vertices $u$ and $v$, and these two vertices are called the {\it endpoints} of edge $e$.

The distance between two vertices $u$ and $v$, denoted by $dist(u,v)$, is the number of edges in a shortest path from $u$ to $v$, and $dist(u,u)=0$ particularly. For a vertex $v$, we define its $i$th level neighborhood as $N_{i}(v)=\{u|dist(u,v)=i\}$, and we denote $N^{k}(v)=\bigcup_{i=1}^{k} N_{i}(v)$.
%Also, we denote $N_{i}[v]=N_{i}[v]\cup \{v\}$ and $N^{k}[v]=N^{k}[v]\cup \{v\}$, which are the closed forms of $N_{i}(v)$ and $N^{k}(v)$ respectively.
The first-level neighborhood $N_1(v)$ is usually denoted as $N(v)$ as well, and we denote $N_{i}[v]=N_{i}(v)\cup \{v\}$.
Also, we define the closed neighborhood of a vertex set S, $N[S] = \bigcup_{v\in S} N[v]$.

A dominating set of $G$ is a subset $D \subseteq V$ such that every vertex in $G$ either belongs to $D$ or is adjacent to a vertex in $D$. The Minimum Dominating Set (MDS) problem calls for finding a dominating set of minimum cardinality. In the Minimum Weight Dominating Set (MWDS) problem, each vertex $v$ is associated with a positive weight $w(v)$, and the task is to find a dominating set $D$ which minimizes the total weight of vertices in $D$ (i.e., min$\sum_{v \in D} w(v)$).
%We call a dominating set $V'$ is a connected dominating set if it has the property that any vertex $v \in V'$ can reach any other vertex $u \in V'$ by a path that stays entirely within $V'$. That is, $V'$ induces a connected subgraph of $G$.
%Our local search algorithm for MWDS maintains a candidate solution $S \subseteq V$ during the search. A vertex $v$ is covered by $S$ if at least one vertex in $N[v]$ is contained in $S$, and is uncovered otherwise. %Formally, we use $S[i]=$1 to denote that $S$ contains some vertex $i$ once.
%The \textbf{$state$} of a vertex $v$ is denoted by $s_v \in \{1,0\}$, such that $s_v=1$ means $v\in S$, and $s_v=0$ means $v\notin S$.

%\subsection{The general framework of local search algorithms for the MWDS problem}
\subsection{Local Search for MWDS}

Local search algorithms perform the search on problem's corresponding search space. The search space is implicitly defined by the way that the algorithm transforms a candidate solution into another. For the MWDS problem, local search algorithms usually maintain a candidate solution $S \subseteq V$ during the search. A vertex $v$ is covered by $S$ if $v\in N[S]$, and is uncovered otherwise. Also, the {\it state} of a vertex $v$ is denoted by $s_{v} \in \{1,0\}$, such that $s_{v}$=1 means a vertex $v$ is covered by a candidate solution $S$, and $s_{v}$=0 means it is uncovered. For a vertex, its {\it age} is defined as the number of steps since the last time it changed its state (being added or removed w.r.t. the maintained candidate solution $S$), and when we say the oldest vertex, we refer
to the one with the minimum {\it age} value.

\SetAlFnt{\small}

\begin{algorithm}[ht]

\caption{The general framework of a local search algorithm} 
$S := InitFunction()$ and $S^{*} := S$\;
    \While{not reach terminate condition}
    {
     \If{$S$ is better than $S^{*}$} {$S^{*} := S$\;}
     $S := MoveNeighbourPoisition(S)$\;
    }
\Return $ S^{*}$\;
\end{algorithm}

%The \textbf{$state$} of a vertex $v$ is denoted by $s_v \in \{1,0\}$, such that $s_v=1$ means $v\in S$, and $s_v=0$ means $v\notin S$.
We first present a general framework of local search in Algorithm 1. As can be seen in this framework, the algorithm consists of two stage: the construction stage and the local search stage.
%Local search algorithms for the MWDS problem usually consist of two stages: the construction stage (Line 1) and the local search stage (Line 2 to 5). 
At first, an initial dominating set is constructed by the greedy initialization process. 
After that, the solution is modified iteratively in the local search procedure, trying to find better solutions.
In the local search stage, if a (redundant) dominating set is found, then the algorithm removes a vertex from $S$ and begins to search for a dominating set with smaller weight, until some stop criterion is reached.
When the current candidate solution is not a dominating set, the move to a neighboring candidate solution consists of two phases: (1) removing a vertex from $S$ and (2) adding some vertices into $S$ until $S$ becomes a dominating set.

\subsection{Configuration Checking}
Local search suffers from the cycling phenomenon, i.e., revisiting a candidate solution that has been recently visited. This phenomenon wastes much computation time of a local search algorithm and more importantly prevents it from escaping from local optima.

To overcome the cycling problem, a number of methods have been proposed. The random walk strategy \cite{ref78} picks the solution component randomly with a certain probability and greedily makes the best possible move with another probability. Random restarting \cite{ref79} is used to restart the search from another starting point of search space. Also, allowing non-improving moves with a probability, as in the Simulating Annealing algorithm, can provide more diversification to the search. Glover proposes the tabu method \cite{ref32}, which has been widely used in local search algorithms \cite{ref36,ref34,ref35}. To prevent the local search to immediately return to a previously visited candidate solution, the tabu method forbids reversing the recent changes, where the strength of forbidding is controlled by a parameter called tabu tenure. Besides these general methods dealing with the cycling phenomenon, there are also heuristics specialized for problems, such as the promising decreasing variable exploitation for the Boolean Satisfiability (SAT) problem \cite{ref42}.

Recently, an interesting strategy called Configuration Checking (CC) \cite{ref37} was proposed to handle the cycling problem in local search. Also, CC does not have instance-dependent parameters and is easy to use. The relationship between the tabu and CC strategy has been thoroughly discussed \cite{ref40}. If the tabu tenure is set to 1, it can be proved that given a variable, if it is forbidden to pick by the tabu method, it is also forbidden by the CC strategy, while its reverse is not necessarily true.

The MWDS problem is in some sense similar to the vertex cover problem in that their tasks are both to find a set of vertices. Thus, we can easily devise a CC strategy for the MWDS problem, following the one for the vertex cover problem \cite{ref37}. An important concept of the CC strategy is the configuration of vertices. Typically, the configuration of a vertex $v$ refers to a vector consisting of the states of all $v$'s neighboring vertices. 
The CC strategy for the MWDS problem can be described as following: given the candidate solution $S$, for a vertex $v\notin S$, if its configuration has not changed since $v$'s last removal from $S$, which means the circumstance of $v$ has not changed, then $v$ should not be added back to $S$.

An implementation of the CC strategy is to apply a Boolean array $confchange$ for vertices, where $confchange(v)$=1 means that $v$ is allowed to be added to $S$, and $confchange(v)$=0 on the contrary. In the beginning, for each vertex $v$, the value of $confchange(v)$ is initialized as 1; afterwards, when removing vertex $x$, $confchange(x)$ is set to 0; whenever a vertex $v$ changes its state, for each vertex $u\in N(v)$, $confchange(u)$ is set to 1.

\section{Two-Level Configuration Checking}
Since the CC strategy has been successfully applied to solve several NP-hard combinatorial optimization problems, a natural question arises whether this strategy can also be applied to MWDS. Unfortunately, a direct application of the original CC strategy in local search for the MWDS problem does not result in an effective algorithm, and has poor performance on a large portion of the benchmark instances.

We observe that, the original CC strategy would mislead the search by forbidding too many candidate vertices. In CC strategy, only the first-level neighborhood of a vertex is considered to avoid cycling, and the configuration of a vertex is considered changed only if at least one of its neighboring vertices changed its state. However, some analysis suggests that not only the first-level neighborhood but the second-level neighborhood are related to the cycling phenomenon and should be considered in the configuration of a vertex.

Inspired by this consideration, we propose a new variant of CC for MWDS, which is referred to as two-level configuration checking (CC$^{2}$ for short), by redefining the configuration of vertices. In the CC$^{2}$ strategy, we consider not only the first-level neighborhood ($N_1$) but also the second-level neighborhood ($N_2$).

\subsection{Definition and Implementation of the CC$^{2}$ Strategy}
In this subsection, we define the CC$^2$ strategy and present an implementation for it. We start from the formal definition of the configuration of a vertex $v$.

\begin{definition}
Given an undirected graph $G=(V,E)$ and $S$ the candidate solution, the {\it configuration} of a vertex $v\in V$ is a vector consisting of state of all vertices in $N^2(v)$.
\end{definition}

Based on the above definition, we can define an important vertex in local search as follows.

\begin{definition}
Given an undirected graph $G=(V,E)$ and $S$ the candidate solution, for a vertex $v\notin S$, $v$ is configuration changed if at least one vertex in $N^2(v)$ has changed its state since the last time $v$ is removed from $S$.
\end{definition}

In the CC$^2$ strategy, only the configuration changed vertices are allowed to be added to the candidate solution $S$.

We implement CC$^{2}$ with a Boolean array $ConfChange$ whose size equals the number of vertices in the input graph. For a vertex $v$, the value of $ConfChange[v]$ is an indicator --- $ConfChange[v]$=1 means $v$ is a configuration changed vertex and is allowed to be added to the candidate solution $S$; otherwise, $ConfChange[v]$=0 and it cannot be added to $S$. During the search procedure, the $ConfChange$ array is maintained as follows.

\textbf{CC$^{2}$-RULE1.} At the start of search process, for each vertex $v$, $ConfChange[v]$ is initialized as 1.

\textbf{CC$^{2}$-RULE2.} When removing a vertex $v$ from the candidate solution $S$, $ConfChange[v]$ is set to 0, and for each vertex $u \in N^{2}(v)$, $ConfChange[u]$ is set to 1.

\textbf{CC$^{2}$-RULE3.} When adding a vertex $v$ into the candidate solution $S$, for each vertex $u \in N^{2}(v)$, $ConfChange[u]$ is set to 1.

To understand RULE2 and RULE3, we note that if $u\in N^{2}(v)$, then $v\in N^{2}(u)$. Thus, if a vertex $v$ changes its state (i.e., either being removed or added w.r.t. the candidate solution), the $Configuration$ of any vertex $u\in N^{2}(v)$ is changed.

\subsection{Relationship between CC and CC$^{2}$ Strategies}
Configuration checking is a general idea, and both the original CC strategy and the CC$^{2}$ strategy are evolved from this idea. Both strategies prefer to pick configuration changed vertices. The difference between these two strategies lies on the concept of configuration changed vertices.

In the following, we compare these two kinds of configuration changed vertices. Let us use $CCV$ to denote the set of configuration changed vertices according to the original CC strategy, and use $CCV^{2}$ to denote the set of configuration changed vertices according to the CC$^{2}$ strategy.

\begin{proposition}
Given an undirected graph $G=(V,E)$ and $S$ the maintained candidate solution, assuming that in some step $t$ we have $CCV=CCV^{2}$, then we can conclude that in the next step, for any $v\in V$, if $v\in CCV$, then $v\in CCV^{2}$.
\end{proposition}

%cai begin here

\textbf{Proof.}
Let us use $v^*$ to denote the vertex selected to change the state in step $t$ of the local search stage for MWDS. In step $t+1$, for any vertex $v$, we have two cases. 1) $v$ was in $CCV$ in step $t$. Since in step $t$ we have $CCV=CCV^{2}$, thus we also have $v\in CCV^2$; 2) $v$ was not in $CCV$ in step $t$. If $v\in CCV$ in step $t+1$, according to the definition of CCV, this can happen only when $v\in N[v^*]$. Because $N[v^*]\subset N^2[v^*]$, we also have $v \in N^2[v^*]$, and thus $v\in CCV^{2}$ in step $t+1$.

The above proposition gives an insight that the size of $CCV^{2}$ is larger than that of $CCV$.
It is easy to see that the reverse of Proposition 1 is not necessarily true.
So, we can easily deduce the following conclusion.

%For any $v\in CCV$, at least one vertex in $N_1(v)$ has changed its state since the last time $v$ was removed from $S$. Let us use $u^*$ to denote a vertex in $N_1(v)$ such that its state has changed since the last time $v$ was removed from $S$. Since $N_1(v)\subset N^2(v)$, then we have $u^*\in N^2(v)$. Therefore, there is at least one vertex in $N^2(v)$ has changed its state since the last time $v$ was removed from $S$, and thus $v$ is a configuration changed variable w.r.t. the CC$^2$ strategy. That is, $v\in CCV^{2}$.

\textbf{Remark 1}
In most conditions, the $CCV^{2}$ set is a superset of $CCV$ set.

%In the CC$^{2}$ strategy, $CCV^{2}$ is adopted as the candidate set of vertices for adding to the candidate solution.
For an algorithm, in each step, there are more candidate vertices which could be added into the candidate solution under the CC$^2$ strategy than under the CC strategy. For this reason, we have more options and thus can explore more search spaces by choosing vertex from $CCV^{2}$ than from $CCV$.

%cai end here

\section{A Novel Scoring Function for MWDS}
Local search algorithms decide which vertex to be selected according to their scores. Thus, the scoring function for vertices plays a critical role in the algorithm, which has direct impact on the intensification and diversification of the search.% How to define the score of vertex affects the diversity of solutions, which plays an important role in local search. And the good diversity score heuristic could escape from local optimal easily.

\subsection{The Previous Scoring Function for MWDS}

Recently, four scoring functions are presented to solve MWDS \cite{ref17}. We list the definitions of these score as below.

$\displaystyle{score_1(u)=\frac{d(u)}{w(u)}}$

$\displaystyle{score_2(u)=\frac{W(u)}{w(u)}}$

$score_3(u)=W(u)-w(u)$

$\displaystyle{score_4(u)=\frac{d(u) \times W(u)}{w(u)}}$

where $u \notin S$, $d(u)$ is the number of uncovered (non-dominated) neighbours of a given vertex $u$ and $W(u)$ is the sum of the weights of the uncovered neighbours of $u$.

The results of this experiment \cite{ref17} show that the first, the second and the last heuristic functions result in a similar performance with no clear pattern of a particular greedy heuristic being better for general graph instances. In contrast, the third greedy heuristic clearly results in the poorest performance.

The weight of vertices being covered in $score_2$ is taken into account \cite{ref16}. In ACO-PP-LS \cite{ref17}, when a newly obtained solution may not be a dominating set, it is repaired by iteratively adding vertices from some vertices not in this obtained solution according to two strategies. With a certain probability, the first strategy is used, otherwise the second strategy is used. Specially, the first strategy is greedy and adds a vertex by using $score_2$, while the second strategy adds a vertex into the candidate solution randomly. ABC \cite{ref18} prunes some redundant vertices according to the greedy strategy $score_1$. In EA/G-IR \cite{ref75}, there are two modifications in $score_2$. The first modification uses the closed neighborhood $N_{1}[u]$ instead of $N_{1}(u)$ when computing $score_2(u)$. The second modification is a tie-breaking rule. Actually, when there are more than one vertex satisfying $score_2$, then the next vertex to be added is selected by this tie-breaking rule. $score_1(u)$ is also computed by the closed neighborhood $N_{1}[u]$ and then the next vertex to be added is determined with this improved $score_1(u)$. In case there is still more than one vertices satisfying $score_1$, then the next vertex to be added is selected arbitrarily from these vertices. In R-PBIG \cite{ref90}, each vertex is rated based on greedy functions $score_1$ or $score_2$ and how to select which greedy function is based on a random number.

\subsection{The Frequency based Scoring Function}
%\question{All previous scoring functions cannot reflect the real-time circumstance of candidate solution, as most of them are based on both static information such as vertex weight and dynamic information such as the current degree and weight of vertex which is calculated according to its uncovered neighbour set.}
%they are based on static information which does not change during the search. %To overcome this problem, we introduce the concept of cost scheme into solving MWDS and propose a dynamic score heuristic.
In this paper, we introduce a novel scoring function by taking into account of the vertices' frequency, which can be viewed as some kind of dynamic information indicating the accumulative effectiveness that the search has on the vertex. Intuitively, if a vertex is usually uncovered, then we should encourage the algorithm to select a vertex to make it covered.

In detail, in a graph, each vertex $v \in V$ has an additional property, frequency, denoted by $freq[v]$. The $freq$ of each vertex is initialized to 1. After each iteration of local search, the $freq$ value of each uncovered vertex is increased by one.
During the search process, we apply the $freq$ of vertex to decide which vertex to be added or removed. Based on this consideration, we propose a new score function, which is formally defined as below.

\begin{definition}
For a graph $G=(V, E)$, and a candidate solution $S$, the frequency based scoring function denoted by $score_{f}$, is a function such that

\begin{eqnarray}
score_{f}(u)=
\left\{
\begin{aligned}
\frac{1}{w(u)}\times\sum_{v \in C_{1} } freq[v], u \notin S,\\
%&\quad when&\quad adding&\quad e&\quad into&\quad CS,\\
-\frac{1}{w(u)}\times\sum_{v \in C_{2} } freq[v], u \in S,
\end{aligned}
\right.
\label{eq1}
\end{eqnarray}
where $C_{1}$=$N[u] \setminus N[S]$ and $C_{2}$=$N[u]$ $\setminus N[S\setminus \{u\}]$.
\end{definition}

Remark that, in the above definition, $C_{1}$ is indeed the set of uncovered vertices that would become covered by adding $u$ into $S$ and $C_{2}$ is the set of covered vertices that would become uncovered by removing $u$ from $S$.

\section{The Selection Vertex Strategy}
During the search process, for preventing visiting previous candidate solutions, we not only use the CC$^{2}$ strategy in the adding process, but also use the forbidding list in the removing process. The $forbid\_list$ used here is a tabu list which keeps track of the vertices added in the last step, and these vertices are prevented from being removed within the tabu tenure. In this sense, this frequency based prohibition mechanism can be viewed as an instantiation of the longer term memory tabu search, and the main difference is that our method also consider the information from the $CC^{2}$ strategy.
 %Beside we use the frequency based scoring function, when finding the removed or added vertex. Firstly we give two remove rules as below.

The algorithm picks a vertex to add or remove, using the frequency based scoring function and the above two strategies. Firstly, we give two rules for removing vertices.

\textbf{REMOVE-RULE1.} Removing one vertex $v$, which has the highest value of $score_{f}(v)$, breaking ties by selecting the oldest one.

\textbf{REMOVE-RULE2.} Removing one vertex $v$, which is not in $forbid\_list$ and has the highest value of $score_{f}(v)$, breaking ties by selecting the oldest one.

When the algorithm finds a solution, it removes one vertex from the solution and continues to search for a solution with smaller weight. In this process, we use REMOVE-RULE1 to pick the vertex.
During the search for a solution, the algorithm exchanges some vertices, i.e., removing one vertex from the candidate solution and then iteratively adding vertices into the candidate solution. In this case, we select one vertex to remove according to REMOVE-RULE2.

The rule to select the adding vertices is given below.

\textbf{ADD-RULE.} Adding one vertex $v$ with $ConfChange[v] \neq 0$, which has the greatest value $score_{f}(v)$, breaking ties by selecting the oldest one.

When adding one vertex into the candidate solution, we try to make the resulting candidate solution's cost (i.e., the total weight of uncovered vertices) as small as possible. When adding one configuration changed vertex with the highest value $score_{f}(v)$, breaking ties by preferring the oldest vertex.

\section{CC$^{2}$FS Algorithm}
Based on CC$^{2}$ and the frequency based scoring function, we develop a local search algorithm named CC$^{2}$FS. During the process of local search, we maintain a set from which the vertex to be added is chosen. The set for finding a vertex to be removed from the candidate solution is simply $S$.

$CCV^{2}=\{v|ConfChange[v] = 1, v \notin S \}$

The pseudo code of CC$^{2}$FS is shown in Algorithm 2. At first, CC$^{2}$FS initializes $ConfChange$, $forbid\_list$ and the $frequency$ and $score_{f}$ of vertices. Then it gets an initial candidate solution $S$ greedily by iteratively adding the vertex that covers the most remaining uncovered vertices until $S$ covers all vertices. At the end of initialization, the best solution $S^{*}$ is updated by $S$.

\SetAlFnt{\small}

\begin{algorithm}[ht]

\caption{CC$^{2}$FS ($G$, {\it cutoff})} \label{alg:main}
\KwIn{a weighted graph $G=(V,E,W)$, the {\it cutoff} time }

\KwOut{dominating set of $G$ }

initialize $ConfChange$, $forbid\_list$, and the $freq$ and $score_{f}$ of of vertices\;
$S := InitGreedyConstruction()$ and $S^{*} := S$\;
    \While{elapsed time $<$ {\it cutoff}}
    {
      \If{there are no uncovered vertices} {
         \lIf{$ w(S)< w(S^{*})$} {$S^{*} := S$}

         $v :=$ a vertex in $S$ with the highest value $score_{f}(v)$, breaking ties in the oldest one\;
         $S := S \setminus \{v\}$ and update $ConfChange$ according to CC$^{2}$-RULE2\;
         continue\;
      }
       $v :=$ a vertex in $S$ with the highest value $score_{f}(v)$ and $v \notin forbid\_list$, breaking ties in the oldest one\;
      $S := S \setminus \{v\}$ and update $ConfChange$ according to CC$^{2}$-RULE2\;
      $forbid\_list := \emptyset $\;
      \While{ there are uncovered vertices}
      {
         $v :=$ a vertex in $CCV^{2}$ with the highest value $score_{f}(v)$, breaking ties in the oldest one\;
        % \lIf{$ w(S)+w(v)\geq w(S^{*})$} {break}
         $S := S \cup \{v\}$ and update $ConfChange$ according to CC$^{2}$-RULE3\;
         $forbid\_list := forbid\_list \cup \{v\} $\;
         $freq[v] := freq[v]+1$, for $v \notin N[S]$\;
      }
    }
\Return $ S^{*}$\;
\end{algorithm}

After initialization, the main loop from lines 3 to 16 begins by checking whether $S$ is a solution (i.e., covers all vertices). When the algorithm finds a better solution, $S^{*}$ is updated. Then one vertex with the highest $score_{f}$ value in $S$ is selected to be removed, breaking tie in favor of the oldest one.
% and the algorithm goes to search for a dominating set of smaller size.
Finally, the values of $ConfChange$ are updated by CC$^{2}$-RULE2.

If there are uncovered vertices, CC$^{2}$FS first picks one vertex to remove from $S$ with the highest value $score_{f}$, breaking tie in favor of the oldest one. Note that when choosing a vertex to remove, we do not consider those vertices in $forbid\_list$, as they are forbidden to be removed by the forbidden list. After removing a vertex, CC$^{2}$FS updates the $ConfChange$ values according to CC$^{2}$-RULE2, and clear $forbid\_list$. Additional, since the tabu tenure is set to be 1, the $forbid\_list$ shall be cleared to allow previous forbidden vertices to be added in subsequent loop.
%The reason why the entire $forbid\_list$ is cleared is that $forbid\_list$ is used to prevent the just previous added vertices from being removed from the candidate solution.

After the removing process, CC$^{2}$FS iteratively adds one vertex into $S$ until it covers all vertices, i.e. the candidate solution is a dominating set. CC$^{2}$FS first selects $v \in CCV^{2}$ with the greatest $score_{f}(v)$, breaking ties in favor of the oldest one. When the picked uncovered vertex is added into the candidate solution, the $ConfChange$ values are updated according to  CC$^{2}$-RULE3 and this added vertex is added into the $forbid\_list$. After adding an uncovered vertex each time, the frequency of uncovered vertices is increased by one. When the time limit reaches, the best solution will be returned.

Now we shall analyse the time complexity of CC$^{2}$FS. For each iteration:
\begin{itemize}
\item The algorithm first dedicates to find a dominate set (Line 4-8). The worst time complexity for finding the removal vertex $v_{i}$ is $O(|S|)$ (Line 6), where $|S|$ denotes the size of the candidate solution $S$. Then, the algorithm updates the value of the related $score_{f}$ as well as $ConfChange$ (Line 10) and the worst time complexity is $O(\Delta(G)^{2})$, %where $|N(v_{max})|$ has the maximum number of $|N(v)|$, for $\forall v \in V$.
where $\Delta(G) = max\{|N[v]||v \in V, G=(V,E)\}$.
\item Otherwise, the algorithm first decides which vertex $v_{j}$ should be deleted and the worst time complexity is also $O(|S|+\Delta(G)^{2})$ (Line 9-11). 
Let $L$ denote the number of uncovered vertices of $S$ (Line 12-16), and under the worst condition the step of the adding procedure is $L=|N[v_{j}]|$.
During this adding procedure, the worst time complexity per step for adding one vertex is $O((|V|-|S|))$ (Line 13). When updating the related $score_{f}$ and $ConfChange$ (Line 14), the worst time complexity per step is also $O(\Delta(G)^{2})$.  
Then, the worst time complexity for the whole adding procedure (Line 12-16) is $O(|N[v_{j}]|(|V|-|S|+\Delta(G)^{2}))$. 
\end{itemize}
Therefore, each iteration of the local search stage of CC$^{2}$FS has a time complexity of $O(|S|+\Delta(G)^{2}+|N[v_{j}]|(|V|-|S|+\Delta(G)^{2}))=O(\Delta(G)^{2}+\Delta(G)(|V|-|S|+\Delta(G)^{2}))=
O(max\{\Delta(G)|V|$, $\Delta(G)^3\})$.

\section{Empirical Results}
We compare CC$^{2}$FS with five competitors on a broad range of benchmarks, with respect to both solution quality and run time. The run time is measured in CPU seconds. Firstly, we introduce the test instances, five competitors and the experimental preliminaries of our experiments.

There are eight benchmarks selected in our experiment, including T1, T2, UDG, two weighted versions of DIMACS, two weighted versions of BHOSLIB, as well as many real world massive graphs. We introduce the benchmarks in the following.

\begin{itemize}
  \item T1 benchmark \cite{ref16} (530 instances): all instances are connected undirected graphs with the vertex weights randomly distributed in the interval [20,70].
  \item T2 benchmark \cite{ref16} (530 instances): the weight of all vertices in all instances is assigned as be a function based on the degree $d(v)$ of the vertex $v$ which is randomly set to in the interval [1, $d(v)^{2}$].
  \item UDG benchmark \cite{ref16} (120 instances): these instances are generated by  using the topology generator \cite{ref50}. Transmission range of all vertices in UDG instances is fixed to either 150 or 200 units.
  \item BHOSLIB benchmark with weighting function $w(v_{i})$=($i$ mod 200)+1 (41 instances): BHOSLIB instances \cite{ref52} were originally unweighted and generated randomly based on the model RB \cite{ref54,ref55,ref53}.
  \item BHOSLIB benchmark with weighting function $w(v_{i})$=1 (41 instances): this benchmark is to test algorithms on uniform weight BHOSLIB graphs.
  \item DIMACS benchmark with weighting function $w(v_{i})$=($i$ mod 200)+1 (54 instances): DIMACS is the most frequently used for comparison and evaluation of graph algorithms. More specifically, the size of the DIMACS instances ranges from less than 150 vertices and 300 edges up to more than 4,000 vertices and 7,900,000 edges. We use the complement graphs of some instances, including the c-fat.clq and p-hat.clq set, to test the efficiency of our algorithm. The original DIMACS graphs are unweighted.
  \item DIMACS benchmark with weighting function $w(v_{i})$=1 (54 instances): this benchmark is to test algorithms on uniform weighted DIMACS graphs.
  \item Massive graph benchmark with weighting function $w(v_{i})$=($i$ mod 200)+1 (74 instances): these were transformed from the unweighted graphs in Network Data Repository online \cite{ref65}.
\end{itemize}

For T1, T2, UDG instances, we note that ten instances are generated for each combination of number of nodes and transmission range. As for the weighting function: for the $i$th vertex $v_{i}$, $w(v_{i})$=($i$ mod 200)+1, it was proposed in the work of Pullan \citeyear{ref51} and has been widely used in the literature for algorithms for solving problems on vertex weighted graphs.

We compare CC$^{2}$FS with HGA \cite{ref17}, ACO-PP-LS \cite{ref17}, ABC \cite{ref18}, EA/G-IR \cite{ref75}, and R-PBIG \cite{ref90}.
%In our paper, we use the settings for the parameter settings introduced in corresponding literatures.
Among them, R-PBIG and ACO-PP-LS are the best available algorithms for solving MWDS.
We have the source code of ACO-PP-LS, so in this paper we use its code to test all benchmarks to get better solution values than those values \cite{ref17}.

We implement CC$^{2}$FS in C++ and compile it by g++ with the -O2 option.
All the experiments are run on Ubuntu Linux, with 3.1 GHZ CPU and 8GB memory.
For T1, T2, and UDG instances, CC$^{2}$FS and ACO-PP-LS are performed once, where one run is terminated upon reaching a given time limit. Among this, the parameter time limit is set to 50 seconds when the number of vertices is less than 500, otherwise the time limit is set to 1000 seconds. We report the real time $RTime$ of ACO-PP-LS and CC$^{2}$FS, while we also give the finial execution time $FTime$ of EA/G-IR and R-PBIG. The real time is a time when ACO-PP-LS and CC$^{2}$FS obtain the best solution respectively. The MEAN contains the average solution values for each of the ten instances of graphs of a particular size.

%We also report the real time $RTime$ of ACO-PP-LS and CC$^{2}$FS, respectively.

For DIMACS, BHOSLIB, and massive graphs instances, our algorithm and ACO-PP-LS are performed 10 independent runs with different random seeds, where each run is terminated upon reaching a given time limit 1000 seconds. The MIN and AVG column contains the minimal and average solution values for each instance by performing 10 runs with different random seeds. The SD column contains the standard deviation for each instance by performing 10 runs. The bold value indicates the best solution value among the different algorithms compared.

Note that for DIMACS, BHOSLIB, and massive graphs benchmarks, we only compare our algorithm with ACO-PP-LS. This is because, 1) as we mentioned, seen from the literatures, R-PBIG and ACO-PP-LS are the best available algorithms for solving MWDS; 2) we have the source code of ACO-PP-LS, while the source code of R-PBIG is not available to us.

\subsection{Results on T1 and T2 Benchmarks}
The performance results of previous algorithms on the T1 benchmark are displayed in Table~\ref{tab_1}. More importantly, this table also summarizes the experimental results on the first benchmark for our algorithm.

%cai begin

Among previous algorithms, for most instances, R-PBIG and ACO-PP-LS can find better solutions than HGA, ABC and EA/G-IR, with only a few exceptions.

For our algorithm, we show the minimum solution value and the run time. As is clear from the Table~\ref{tab_1}, CC$^2$FS shows significant superiority on the T1 benchmark, except v50e750. By comparing these algorithms, we can easily conclude that CC$^{2}$FS outperforms other algorithms.

The experimental results on the T2 benchmark are presented in Table~\ref{tab_2}. The quality of the solutions found by CC$^{2}$FS is always much smaller than those found by other algorithms on all instances with 2 exceptions, i.e. v250e250 and v800e10000.

%\question{
%In \cite{ref90}, a hybrid algorithmic model is proposed, named Hyb-R-PBIG in which the first half of the time limit is spent by R-PBIG, and the second half is allocated to the mathematical programming solver CPLEX which is executed in the solution polishing mode applied to the best solution obtained by R-PBIG.
%Our algorithm CC$^{2}$FS can be also combined with CPLEX, resulting in a new hybrid algorithm which reaches the same or better solution values than Hyb-R-PBIG.}

%cai end

\begin{table}[tbp]
\centering
\renewcommand{\arraystretch}{1}
\caption{Experiment results of HGA, ACO-PP-LS, ABC, EA/G-IR, R-PBIG, and CC$^{2}$FS on the T1 benchmark. {\small Each set contains 10 instances.}}
%\begin{tabular}{p{2cm}|p{0.5cm}p{0.5cm}p{0.5cm}p{0.5cm}p{0.5cm}p{1.5cm}}
\scriptsize\setlength{\tabcolsep}{3pt}
\begin{tabular}{|l|l|lr|l|lr|lr|lr|}
\hline
\multicolumn{1}{|c|}{Instance} & \multicolumn{1}{c|}{HGA} & \multicolumn{2}{c|}{ACO-PP-LS} & \multicolumn{1}{c|}{ABC} & \multicolumn{2}{c|}{EA/G-IR} & \multicolumn{2}{c|}{R-PBIG} & \multicolumn{2}{c|}{CC2FS}      \\
\multicolumn{1}{|c|}{T1}       & MEAN                     & MEAN               & RTime     & MEAN                     & MEAN              & FTime    & MEAN              & FTime   & MEAN            & RTime         \\ \hline
v50e50                         & \textbf{531.3}           & \textbf{531.3}     & 0.2       & 534                      & 532.9             & 0.21     & \textbf{531.3}    & 0.5     & \textbf{531.3}  & \textless0.01 \\
v50e100                        & 371.2                    & 371.2              & 0.17      & 371.2                    & 371.5             & 0.23     & 371.1             & 0.8     & \textbf{370.9}  & \textless0.01 \\
v50e250                        & \textbf{175.7}           & \textbf{175.7}     & 0.12      & \textbf{175.7}           & \textbf{175.7}    & 0.18     & \textbf{175.7}    & 1.3     & \textbf{175.7}  & \textless0.01 \\
v50e500                        & \textbf{94.9}            & \textbf{94.9}      & 0.05      & \textbf{94.9}            & \textbf{94.9}     & 0.16     & 95                & 2.3     & \textbf{94.9}   & \textless0.01 \\
v50e750                        & \textbf{63.1}            & \textbf{63.1}      & 0.03      & \textbf{63.1}            & 63.3              & 0.1      & 63.8              & 2.5     & 63.3            & \textless0.01 \\
v50e1000                       & \textbf{41.5}            & \textbf{41.5}      & 0.01      & \textbf{41.5}            & \textbf{41.5}     & 0.1      & \textbf{41.5}     & 3       & \textbf{41.5}   & \textless0.01 \\
v100e100                       & 1081.3                   & 1065.6             & 2.05      & 1077.7                   & 1065.5            & 0.76     & 1061.9            & 1.4     & \textbf{1061}   & 0.04          \\
v100e250                       & 626.2                    & 623.1              & 1.3       & 621.6                    & 620               & 0.72     & 619.3             & 2.2     & \textbf{618.9}  & 0.04          \\
v100e500                       & 358.3                    & 360.6              & 0.54      & 356.4                    & 355.9             & 0.6      & 356.5             & 3       & \textbf{355.6}  & \textless0.01 \\
v100e750                       & 261.2                    & 261                & 0.46      & 255.9                    & 256.7             & 0.52     & 256.5             & 3.7     & \textbf{255.8}  & \textless0.01 \\
v100e1000                      & 205.6                    & 207.3              & 0.38      & \textbf{203.6}           & \textbf{203.6}    & 0.49     & \textbf{203.6}    & 4.3     & \textbf{203.6}  & \textless0.01 \\
v100e2000                      & 108.2                    & 108.4              & 0.2       & 108.2                    & 108.1             & 0.45     & 108               & 6.2     & \textbf{107.4}  & 0.19          \\
v150e150                       & 1607                     & 1582               & 4.88      & 1607.9                   & 1587.4            & 1.64     & 1582.5            & 2.4     & \textbf{1580.5} & 0.02          \\
v150e250                       & 1238.6                   & 1228.4             & 3.67      & 1231.2                   & 1224.5            & 1.71     & 1219.5            & 3.3     & \textbf{1218.2} & 0.06          \\
v150e500                       & 763                      & 763                & 2.2       & 752.1                    & 755.3             & 1.45     & 745               & 4.3     & \textbf{744.6}  & 0.05          \\
v150e750                       & 558.5                    & 554                & 1.46      & 549.3                    & 550.8             & 1.27     & 548.2             & 5.2     & \textbf{546.1}  & 0.04          \\
v150e1000                      & 438.7                    & 440.7              & 1.34      & 435.1                    & 435.2             & 1.11     & 433.6             & 5.9     & \textbf{432.9}  & 0.03          \\
v150e2000                      & 245.7                    & 251.8              & 0.89      & 242.2                    & 241.5             & 0.88     & 241.5             & 8.7     & \textbf{240.8}  & 0.17          \\
v150e3000                      & 169.2                    & 171.4              & 0.8       & 167.8                    & 168.1             & 0.82     & 168.4             & 11.1    & \textbf{166.9}  & 0.06          \\
v200e250                       & 1962.1                   & 1919.5             & 9.54      & 1941.1                   & 1924.1            & 3.68     & 1914.6            & 4.7     & \textbf{1910.4} & 0.19          \\
v200e500                       & 1266.3                   & 1252.9             & 4.92      & 1246.9                   & 1251.3            & 3.3      & 1235.3            & 6.2     & \textbf{1232.8} & 1.01          \\
v200e750                       & 939.8                    & 934.3              & 3.71      & 923.7                    & 927.3             & 2.78     & 914.9             & 7.4     & \textbf{911.2}  & 0.45          \\
v200e1000                      & 747.8                    & 741.8              & 2.99      & 730.4                    & 731.1             & 2.39     & 725.2             & 8.2     & \textbf{724}    & 0.25          \\
v200e2000                      & 432.9                    & 437.3              & 1.36      & 417.6                    & 417               & 1.68     & 414.8             & 10.9    & \textbf{412.7}  & 0.45          \\
v200e3000                      & 308.5                    & 308.8              & 1.24      & 294.4                    & 294.7             & 1.42     & 294.2             & 14.2    & \textbf{292.8}  & 0.41          \\
v250e250                       & 2703.4                   & 2646.6             & 16.09     & 2685.7                   & 2653.7            & 4.65     & 2653.7            & 6       & \textbf{2633.4} & 0.2           \\
v250e500                       & 1878.8                   & 1840.1             & 10.91     & 1836                     & 1853.3            & 4.66     & 1812.6            & 8.6     & \textbf{1805.9} & 0.92          \\
v250e750                       & 1421.1                   & 1396.8             & 7.15      & 1391.9                   & 1399.2            & 4.25     & 1368.6            & 9.6     & \textbf{1362.2} & 0.65          \\
v250e1000                      & 1143.4                   & 1120.2             & 5.53      & 1115.3                   & 1114.9            & 3.69     & 1097.1            & 10.9    & \textbf{1091.1} & 0.48          \\
v250e2000                      & 656.6                    & 666                & 3.23      & 630.5                    & 637.5             & 2.65     & 624.7             & 14      & \textbf{621.9}  & 0.44          \\
v250e3000                      & 469.3                    & 469.4              & 2.64      & 454.9                    & 456.3             & 2.16     & 451.5             & 17.9    & \textbf{447.9}  & 0.72          \\
v250e5000                      & 300.5                    & 307                & 2.29      & 292.4                    & 291.8             & 5.16     & 291.5             & 25.4    & \textbf{289.5}  & 0.17          \\
v300e300                       & 3255.2                   & 3190.6             & 24.39     & 3240.7                   & 3213.7            & 8.73     & 3189.3            & 7.6     & \textbf{3178.6} & 1.47          \\
v300e500                       & 2509.8                   & 2461.4             & 21        & 2484.6                   & 2474.8            & 7.21     & 2446.9            & 10.2    & \textbf{2438.1} & 1.46          \\
v300e750                       & 1933.9                   & 1885.1             & 16.91     & 1901.4                   & 1896.3            & 6.48     & 1869.6            & 12      & \textbf{1854.6} & 1.6           \\
v300e1000                      & 1560.1                   & 1532.7             & 10.49     & 1523.4                   & 1531              & 5.7      & 1503.4            & 13.2    & \textbf{1495}   & 0.61          \\
v300e2000                      & 909.6                    & 900.5              & 5.95      & 875.5                    & 880.1             & 4.03     & 872.5             & 16.8    & \textbf{862.5}  & 1.93          \\
v300e3000                      & 654.9                    & 658.8              & 4.03      & 635.3                    & 638.2             & 3.27     & 629               & 21.3    & \textbf{624.3}  & 1.12          \\
v300e5000                      & 428.3                    & 432.3              & 3.67      & 411                      & 415.7             & 2.59     & 409.4             & 29.8    & \textbf{406.1}  & 1.72          \\
v500e500                       & 5498.3                   & 5370.4             & 99.09     & 5480.1                   & 5380.1            & 37.52    & 5378.4            & 21.1    & \textbf{5305.7} & 2.63          \\
v500e1000                      & 3798.6                   & 3675.8             & 64.96     & 3707.6                   & 3695.2            & 26.36    & 3642.2            & 29.1    & \textbf{3607.8} & 4.24          \\
v500e2000                      & 2338.2                   & 2236.2             & 32.85     & 2266.1                   & 2264.3            & 25.54    & 2203.9            & 36.1    & \textbf{2181}   & 4.83          \\
v500e5000                      & 1122.7                   & 1105.8             & 15.57     & 1070.9                   & 1083.5            & 9.02     & 1055.9            & 55.9    & \textbf{1043.3} & 5.07          \\
v500e10000                     & 641.1                    & 640.9              & 10.57     & 596                      & 606.8             & 6.08     & 596.3             & 76.2    & \textbf{587.2}  & 6.19          \\
v800e1000                      & 8017.7                   & 7991.6             & 174.62    & 7907.3                   & 7792.2            & 129.82   & 7768.6            & 67.9    & \textbf{7663.4} & 10.58         \\
v800e2000                      & 5317.7                   & 5298.4             & 95.34     & 5193.2                   & 5160.7            & 102.09   & 5037.9            & 83.3    & \textbf{4982.1} & 8.83          \\
v800e5000                      & 2633.4                   & 2578.8             & 59.23     & 2548.6                   & 2561.9            & 53.02    & 2465.4            & 122.6   & \textbf{2441.2} & 6.58          \\
v800e10000                     & 1547.7                   & 1512.7             & 41.86     & 1471.7                   & 1497              & 31.17    & 1420              & 171.1   & \textbf{1395.6} & 6.84          \\
v1000e1000                     & 11095.2                  & 10984.9            & 412.14    & 10992.4                  & 10771.7           & 249.82   & 10825             & 96.9    & \textbf{10585.3}  & 12.18         \\
v1000e5000                     & 3996.6                   & 3977.7             & 91.07     & 3853.7                   & 3876.3            & 107.41   & 3693.1            & 184.2   & \textbf{3671.8} & 8.49          \\
v1000e10000                    & 2334.7                   & 2291.8             & 83.82     & 2215.9                   & 2265.1            & 63.22    & 2140.3            & 254.9   & \textbf{2109}   & 9.43          \\
v1000e15000                    & 1687.5                   & 1647.4             & 63.15     & 1603.2                   & 1629.4            & 45.86    & 1549.1            & 282     & \textbf{1521.5} & 11.91         \\
v1000e20000                    & 1337.2                   & 1297.5             & 44.21     & 1259.5                   & 1299.9            & 36.35    & 1219              & 289.8   & \textbf{1203.6} & 11.4          \\ \hline
\end{tabular}
\label{tab_1}
\end{table}

\begin{table}[tbp]
\centering
\renewcommand{\arraystretch}{1}
\caption{Experiment results of HGA, ACO-PP-LS, ABC, EA/G-IR, R-PBIG, and CC$^{2}$FS on the T2 benchmark. {\small Each set contains 10 instances.}}
%\begin{tabular}{p{2cm}|p{0.5cm}p{0.5cm}p{0.5cm}p{0.5cm}p{0.5cm}p{1.5cm}}
\scriptsize\setlength{\tabcolsep}{3pt}
\begin{tabular}{|l|l|lr|l|lr|lr|lr|}
\hline
\multicolumn{1}{|c|}{Instance} & \multicolumn{1}{c|}{HGA} & \multicolumn{2}{c|}{ACO-PP-LS} & \multicolumn{1}{c|}{ABC} & \multicolumn{2}{c|}{EA/G-IR} & \multicolumn{2}{c|}{R-PBIG} & \multicolumn{2}{c|}{CC$^{2}$FS}      \\
\multicolumn{1}{|c|}{T2}       & MEAN                     & MEAN               & RTime     & MEAN                     & MEAN              & FTime    & MEAN              & FTime   & MEAN            & RTime         \\ \hline
v50e50                         & \textbf{60.8}            & \textbf{60.8}      & 0.08      & \textbf{60.8}            & 60.8              & 0.19     & \textbf{60.8}     & 0.5     & \textbf{60.8}   & \textless0.01 \\
v50e100                        & \textbf{90.3}            & \textbf{90.3}      & 0.17      & \textbf{90.3}            & 90.3              & 0.27     & \textbf{90.3}     & 0.8     & \textbf{90.3}   & \textless0.01 \\
v50e250                        & \textbf{146.7}           & \textbf{146.7}     & 0.09      & \textbf{146.7}           & 146.7             & 0.23     & \textbf{146.7}    & 1.4     & \textbf{146.7}  & \textless0.01 \\
v50e500                        & \textbf{179.9}           & \textbf{179.9}     & 0.04      & \textbf{179.9}           & 179.9             & 0.09     & \textbf{179.9}    & 2.1     & \textbf{179.9}  & \textless0.01 \\
v50e750                        & \textbf{171.1}           & \textbf{171.1}     & 0.01      & \textbf{171.1}           & 171.1             & 0.07     & \textbf{171.1}    & 2.4     & \textbf{171.1}  & \textless0.01 \\
v50e1000                       & \textbf{146.5}           & \textbf{146.5}     & 0.01      & \textbf{146.5}           & 146.5             & 0.06     & \textbf{146.5}    & 2.9     & \textbf{146.5}  & \textless0.01 \\
v100e100                       & 124.5                    & \textbf{123.5}     & 1.05      & 124.4                    & 123.5             & 0.6      & \textbf{123.5}    & 1.2     & \textbf{123.5}  & \textless0.01 \\
v100e250                       & 211.4                    & 210.1              & 0.89      & 209.6                    & \textbf{209.2}    & 0.92     & \textbf{209.2}    & 2.1     & \textbf{209.2}  & \textless0.01 \\
v100e500                       & 306                      & \textbf{305.7}     & 0.57      & 305.8                    & \textbf{305.7}    & 0.78     & \textbf{305.7}    & 2.9     & \textbf{305.7}  & \textless0.01 \\
v100e750                       & 385.3                    & \textbf{384.5}     & 0.45      & \textbf{384.5}           & \textbf{384.5}    & 0.7      & 386.9             & 3.5     & \textbf{384.5}  & \textless0.01 \\
v100e1000                      & 429.1                    & 427.7              & 0.21      & \textbf{427.3}           & \textbf{427.3}    & 0.67     & \textbf{427.3}    & 4.2     & \textbf{427.3}  & \textless0.01 \\
v100e2000                      & \textbf{550.6}           & \textbf{550.6}     & 0.15      & \textbf{550.6}           & \textbf{550.6}    & 0.54     & 552.7             & 6.3     & \textbf{550.6}  & \textless0.01 \\
v150e150                       & 186                      & \textbf{184.5}     & 2.9       & 185.9                    & \textbf{184.5}    & 1.85     & \textbf{184.5}    & 2.1     & \textbf{184.5}  & 0.07          \\
v150e250                       & 234.9                    & 233                & 2.7       & 233.4                    & 232.8             & 2.03     & \textbf{232.8}    & 3.1     & \textbf{232.8}  & \textless0.01 \\
v150e500                       & 350                      & 350.3              & 1.49      & \textbf{349.5}           & 349.7             & 1.95     & 349.7             & 4.4     & \textbf{349.5}  & \textless0.01 \\
v150e750                       & 455.8                    & 453                & 1.81      & 453.7                    & \textbf{452.4}    & 1.78     & \textbf{452.4}    & 5.4     & \textbf{452.4}  & \textless0.01 \\
v150e1000                      & 547.5                    & 549                & 1.3       & 547.8                    & 548.2             & 1.61     & 547.8             & 6       & \textbf{547.2}  & \textless0.01 \\
v150e2000                      & \textbf{720.1}           & 720.8              & 0.88      & \textbf{720.1}           & \textbf{720.1}    & 1.2      & \textbf{720.1}    & 8.4     & \textbf{720.1}  & \textless0.01 \\
v150e3000                      & 792.6                    & \textbf{792.4}     & 0.56      & 793.2                    & \textbf{792.4}    & 1.07     & 793.2             & 11.7    & \textbf{792.4}  & 0.66          \\
v200e250                       & 275.1                    & 272.2              & 5.01      & 273.5                    & 272.3             & 4.38     & \textbf{271.7}    & 4.3     & \textbf{271.7}  & \textless0.01 \\
v200e500                       & 390.7                    & 387.4              & 3.71      & 387.6                    & 388.4             & 4.51     & 386.8             & 6.1     & \textbf{386.7}  & 0.04          \\
v200e750                       & 507                      & 499.7              & 3.56      & 498.5                    & 497.2             & 4.18     & \textbf{497.1}    & 7.2     & \textbf{497.1}  & \textless0.01 \\
v200e1000                      & 601.1                    & 598.9              & 2.69      & 599.3                    & 598.2             & 3.89     & \textbf{596.8}    & 8.5     & \textbf{596.8}  & 0.04          \\
v200e2000                      & 893.5                    & 887.3              & 1.88      & 885.5                    & 885.8             & 2.78     & \textbf{884.6}    & 11.4    & \textbf{884.6}  & 0.09          \\
v200e3000                      & 1021.3                   & 1027               & 1.01      & 1021.3                   & 1019.7            & 2.16     & \textbf{1019.2}   & 14.1    & \textbf{1019.2} & 0.06          \\
v250e250                       & 310.1                    & 306.5              & 8.86      & 308.6                    & 306.5             & 5.26     & \textbf{306}      & 4.9     & 306.1           & 0.01          \\
v250e500                       & 444                      & 441.9              & 9.11      & 442.6                    & 441.6             & 6.1      & 441               & 8.2     & \textbf{440.7}  & 0.16          \\
v250e750                       & 578.2                    & 571.4              & 7.64      & 569.9                    & 569.2             & 6.09     & 567.9             & 10      & \textbf{567.4}  & 0.2           \\
v250e1000                      & 672.8                    & 671.5              & 4.81      & 670.3                    & 671.7             & 5.89     & 669.2             & 11.4    & \textbf{668.6}  & 0.17          \\
v250e2000                      & 1030.8                   & 1018.9             & 3.88      & 1010.4                   & 1010.3            & 4.23     & 1009.5            & 14.5    & \textbf{1007}   & 0.48          \\
v250e3000                      & 1262                     & 1261.2             & 2.75      & 1251.3                   & \textbf{1250.6}   & 3.5      & 1251.6            & 18.1    & \textbf{1250.6} & 0.57          \\
v250e5000                      & 1480.9                   & 1469.6             & 1.36      & 1464.7                   & \textbf{1464.2}   & 2.59     & \textbf{1464.2}   & 25.5    & \textbf{1464.2} & 0.01          \\
v300e300                       & 375.6                    & 371.1              & 14.46     & 373.5                    & 370.5             & 9.01     & \textbf{369.9}    & 6.3     & \textbf{369.9}  & 0.13          \\
v300e500                       & 484.2                    & 479.9              & 11.93     & 481.6                    & 480               & 8.83     & 478               & 9.6     & \textbf{477.8}  & 0.06          \\
v300e750                       & 623.8                    & 616.1              & 13.63     & 617.6                    & 613.8             & 7.57     & 613.6             & 11.7    & \textbf{613.3}  & 0.37          \\
v300e1000                      & 751.1                    & 740.9              & 11.37     & 743.6                    & 742.2             & 8.96     & 738.3             & 13.5    & \textbf{737.9}  & 0.28          \\
v300e2000                      & 1106.7                   & 1104.5             & 6.86      & 1095.9                   & 1094.9            & 6.67     & 1094.6            & 17.4    & \textbf{1093.8} & 0.03          \\
v300e3000                      & 1382.1                   & 1398.4             & 6.25      & 1361.7                   & 1359.5            & 5.41     & \textbf{1358.5}   & 20.9    & \textbf{1358.5} & 0.08          \\
v300e5000                      & 1686.3                   & 1691.5             & 3.21      & \textbf{1682.7}          & 1683.6            & 4.02     & 1683.2            & 29.5    & \textbf{1682.7} & 0.01          \\
v500e500                       & 632.9                    & 627.3              & 33.4      & 630.4                    & 625.8             & 31.06    & 624.2             & 17.7    & \textbf{623.6}  & 0.29          \\
v500e1000                      & 919.2                    & 907.6              & 70.92     & 906.7                    & 906               & 28.27    & 901.3             & 28.1    & \textbf{899.8}  & 2.08          \\
v500e2000                      & 1398.2                   & 1381.5             & 38.78     & 1383.6                   & 1376.7            & 23.41    & 1364.4            & 37.2    & \textbf{1363.3} & 2.28          \\
v500e5000                      & 2393.2                   & 2406.9             & 11.87     & 2337.9                   & 2340.3            & 17.36    & 2341.5            & 59      & \textbf{2333.7} & 0.31          \\
v500e10000                     & 3264.9                   & 3277.9             & 6.48      & \textbf{3211.5}          & 3216.4            & 10.8     & 3216.1            & 80.5    & \textbf{3211.5} & 0.06          \\
v800e1000                      & 1128.2                   & 1121.7             & 274.35    & 1119.2                   & 1107.9            & 132.36   & 1107.6            & 59.6    & \textbf{1104.3} & 2.36          \\
v800e2000                      & 1679.2                   & 1674.9             & 97.55     & 1656.4                   & 1641.7            & 111.84   & 1634.6            & 83.5    & \textbf{1632.3} & 3.59          \\
v800e5000                      & 3003.6                   & 3065.7             & 47.02     & 2917.4                   & 2939.3            & 68.14    & 2884.8            & 128.3   & \textbf{2878.5} & 3.65          \\
v800e10000                     & 4268.1                   & 4357.1             & 26.77     & 4121.3                   & 4155.1            & 40.15    & \textbf{4103.7}   & 183.9   & 4105.6          & 1.55          \\
v1000e1000                     & 1265.2                   & 1254.4             & 564.71    & 1256.2                   & 1240.8            & 202.08   & 1243.6            & 80.9    & \textbf{1237.7} & 0.86          \\
v1000e5000                     & 3320.1                   & 3371.6             & 95.9      & 3240.7                   & 3222              & 132.94   & 3195.7            & 196     & \textbf{3178.7} & 8.87          \\
v1000e10000                    & 4947.5                   & 5041.6             & 55.26     & 4781.2                   & 4798.6            & 84.82    & 4722.4            & 274.6   & \textbf{4711.8} & 4.06          \\
v1000e15000                    & 6267.6                   & 6336.1             & 46.07     & 5931                     & 5958.1            & 61.64    & 5884.2            & 305.2   & \textbf{5874.2} & 2.97          \\
v1000e20000                    & 7088.5                   & 7166.7             & 37.65     & 6729                     & 6775.8            & 59.2     & 6678              & 319.2   & \textbf{6662.1} & 2.68          \\ \hline
\end{tabular}
\label{tab_2}
\end{table}

\subsection{Results on UDG Benchmark}
Table~\ref{tab_3} shows the comparative results on the UDG benchmark. For these instances, the solution value obtained by EA/G-IR and ACO-PP-LS almost matches CC$^{2}$FS, expert for one instance V1000U150. But, the real run time of all instances solved by CC$^{2}$FS is always less than 0.2 seconds and thus CC$^{2}$FS solves faster than EA/G-IR and ACO-PP-LS.

Observed from Table~\ref{tab_1}, \ref{tab_2} and \ref{tab_3}, our algorithm uses less time to get better values, while the run time of ACO-PP-LS grows quickly with increasing the number of vertices. Specially, for some big graphs, such as v1000e1000 and V1000U200, the run time of our algorithm is two orders of magnitudes less than of ACO-PP-LS.

\begin{table}[tbp]
\centering
\renewcommand{\arraystretch}{1}
\caption{Experiment results of HGA, ACO-PP-LS, ABC, EA/G-IR, and CC$^{2}$FS on the UDG benchmark. {\small Each set contains 10 instances.}}
%
\label{tab_3}
\end{table}

\begin{table}[tbp]
\centering
\renewcommand{\arraystretch}{1}
\caption{Experiment results of ACO-PP-LS and CC$^{2}$FS on the BHOSLIB benchmark. {\small For the ith vertex $v_{i}$, $w(v_{i})$=($i$ mod 200)+1}}
\scriptsize\setlength{\tabcolsep}{8pt}
\begin{tabular}{|l|llll|llll|}
\hline
\multicolumn{1}{|c|}{\multirow{2}{*}{Instance}} & \multicolumn{4}{c|}{ACO-PP-LS}                                                                          & \multicolumn{4}{c|}{CC$^{2}$FS}                                                                               \\
\multicolumn{1}{|c|}{}                          & \multicolumn{1}{c}{MIN} & \multicolumn{1}{c}{AVG} & \multicolumn{1}{c}{SD} & \multicolumn{1}{c|}{RTime} & \multicolumn{1}{c}{MIN} & \multicolumn{1}{c}{AVG} & \multicolumn{1}{c}{SD} & \multicolumn{1}{c|}{RTime} \\ \hline

frb30-15-1                                      & 223                     & 223.5                   & 1.5                    & 12.16                      & \textbf{214}            & 214                     & 0                      & 0.06                       \\
frb30-15-2                                      & 244                     & 244                     & 0                      & 7.25                       & \textbf{242}            & 242                     & 0                      & 0.02                       \\
frb30-15-3                                      & \textbf{175}            & \textbf{175}                       & 0                      & 5.97                       & \textbf{175}            & \textbf{175}                       & 0                      & \textbf{0.01}              \\
frb30-15-4                                      & 174                     & 182.7                   & 3.77                   & 10.34                      & \textbf{166}            & 167                     & 1.73                   & 0.16                       \\
frb30-15-5                                      & 172                     & 177.4                   & 4.88                   & 11.2                       & \textbf{160}            & 160                     & 0                      & 0.00                       \\
\hline

\end{tabular}
\label{tab_4}
\end{table}

\begin{table}[htbp]
\centering
\caption{Experiment results of ACO-PP-LS and CC$^{2}$FS on the BHOSLIB benchmark. {\small For the $i$th vertex $v_{i}$, $w(v_{i})$=1}}
\scriptsize\setlength{\tabcolsep}{8pt}
\begin{tabular}{|l|llll|llll|}
\hline
\multicolumn{1}{|c|}{\multirow{2}{*}{INSTANCE}} & \multicolumn{4}{c|}{ACO-PP-LS}                                                                          & \multicolumn{4}{c|}{CC$^{2}$FS}                                                                               \\
\multicolumn{1}{|c|}{}                          & \multicolumn{1}{c}{MIN} & \multicolumn{1}{c}{AVG} & \multicolumn{1}{c}{SD} & \multicolumn{1}{c|}{Rtime} & \multicolumn{1}{c}{MIN} & \multicolumn{1}{c}{AVG} & \multicolumn{1}{c}{SD} & \multicolumn{1}{c|}{Rtime} \\ \hline
frb30-15-1                                      & 12                      & 12                      & 0                      & 9.7                        & \textbf{11}             & 11                      & 0                      & 0.09                       \\
frb30-15-2                                      & \textbf{11}             & 11.1                    & 0.3                    & 8.14                       & \textbf{11}             & \textbf{11}                      & 0                      & 0.08              \\
frb30-15-3                                      & 12                      & 12                      & 0                      & 4.4                        & \textbf{10}             & 10                      & 0                      & 2.00                       \\
frb30-15-4                                      & 12                      & 12                      & 0                      & 0.06                       & \textbf{11}             & 11                      & 0                      & 0.11                       \\
frb30-15-5                                      & 12                      & 12.5                    & 0.5                    & 10.3                       & \textbf{11}             & 11                      & 0                      & 0.28                       \\
\hline
\end{tabular}
\label{tab_5}
\end{table}

\begin{table}[tbp]
\centering
\renewcommand{\arraystretch}{1}
\caption{Experiment results of ACO-PP-LS and CC$^{2}$FS on the DIMACS benchmark. {\small For the ith vertex $v_{i}$, $w(v_{i})$=($i$ mod 200)+1}}
\scriptsize\setlength{\tabcolsep}{3pt}\begin{tabular}{|l|llll|llll|}
\hline
\multicolumn{1}{|c|}{\multirow{2}{*}{Instance}} & \multicolumn{4}{c|}{ACO-PP-LS}                                                                          & \multicolumn{4}{c|}{CC$^{2}$FS}                                                                               \\
\multicolumn{1}{|c|}{}                          & \multicolumn{1}{c}{MIN} & \multicolumn{1}{c}{AVG} & \multicolumn{1}{c}{SD} & \multicolumn{1}{c|}{RTime} & \multicolumn{1}{c}{MIN} & \multicolumn{1}{c}{AVG} & \multicolumn{1}{c}{SD} & \multicolumn{1}{c|}{RTime} \\ \hline

brock200\_2                                     & \textbf{23}             & \textbf{23}                     & 0                       & 0.01                      & \textbf{23}             & \textbf{23} & 0                      & \textbf{\textless0.01}     \\
brock200\_4                                     & \textbf{68}             & 70.4                    & 1.2                     & 2.95                      & \textbf{68}             & \textbf{68}                    & 0                      & \textless0.01     \\
brock400\_2                                     & \textbf{65}             & \textbf{65}                                  & 0                       & 3.04                      & \textbf{65}             & \textbf{65}                                  & 0                      & \textbf{\textless0.01}     \\
brock400\_4                                     & \textbf{75}             & 75.7                    & 0.9                     & 5.35                      & \textbf{75}             & \textbf{75}                      & 0                      & 0.01              \\
brock800\_2                                     & \textbf{28}             & 28.4                    & 0.49                    & 25.7                      & \textbf{28}             & \textbf{28}                      & 0                      & \textless0.01     \\
\hline

\end{tabular}
\label{tab_6}
\end{table}

\begin{table}[htbp]

\centering
\caption{Experiment results of ACO-PP-LS and CC$^{2}$FS on the DIMACS  benchmark. {\small For the $i$th vertex $v_{i}$, $w(v_{i})$=1}}

\scriptsize\setlength{\tabcolsep}{1.5pt}
\begin{tabular}{|l|llll|llll|}
\hline
\multicolumn{1}{|c|}{\multirow{2}{*}{INSTANCE}} & \multicolumn{4}{c|}{ACO-PP-LS}                                                                          & \multicolumn{4}{c|}{CC$^{2}$FS}                                                                               \\
\multicolumn{1}{|c|}{}                          & \multicolumn{1}{c}{MIN} & \multicolumn{1}{c}{AVG} & \multicolumn{1}{c}{SD} & \multicolumn{1}{c|}{Rtime} & \multicolumn{1}{c}{MIN} & \multicolumn{1}{c}{AVG} & \multicolumn{1}{c}{SD} & \multicolumn{1}{c|}{Rtime} \\ \hline
brock200\_2                                     & \textbf{4}              & \textbf{4}                       & 0                      & 0.02                       & \textbf{4}              & \textbf{4}                       & 0                      & \textbf{\textless0.01}     \\
brock200\_4                                     & 6                       & 6                       & 0                      & 0.02                       & \textbf{5}              & 5                       & 0                      & 0.04                       \\
brock400\_2                                     & \textbf{9}              & 9.7                     & 0.46                   & 0.08                       & \textbf{9}              & \textbf{9}                        & 0                      & 0.07              \\
brock400\_4                                     & \textbf{9}              & 9.8                     & 0.4                    & 0.03                       & \textbf{9}              & \textbf{9}                        & 0                      & 0.06              \\
brock800\_2                                     & \textbf{8}              & \textbf{8}                       & 0                      & 0.06                       & \textbf{8}              & \textbf{8}                       & 0                      & \textbf{\textless0.01}     \\
\hline
\end{tabular}
\label{tab_7}
\end{table}

\begin{table}[htbp]
\centering
\renewcommand{\arraystretch}{1}
\caption{Experiment results of ACO-PP-LS and CC$^{2}$FS on the massive graph. {\small For the ith vertex $v_{i}$, $w(v_{i})$=($i$ mod 200)+1}}
\scriptsize\setlength{\tabcolsep}{4pt}
\begin{tabular}{|l|ll|llll|llll|}
\hline
\multicolumn{1}{|c|}{\multirow{2}{*}{INSTANCE}} & \multicolumn{1}{c}{\multirow{2}{*}{|V|}} & \multicolumn{1}{c|}{\multirow{2}{*}{|E|}} & \multicolumn{4}{c|}{ACO-PP-LS}                                                                          & \multicolumn{4}{c|}{CC$^{2}$FS}                                                                               \\
\multicolumn{1}{|c|}{}                          & \multicolumn{1}{c}{}                     & \multicolumn{1}{c|}{}                     & \multicolumn{1}{c}{MIN} & \multicolumn{1}{c}{AVG} & \multicolumn{1}{c}{SD} & \multicolumn{1}{c|}{Rtime} & \multicolumn{1}{c}{MIN} & \multicolumn{1}{c}{AVG} & \multicolumn{1}{c}{SD} & \multicolumn{1}{c|}{Rtime} \\ \hline
bio-dmela           & 7393   & 25569    & 117222    & 117222  & 0      & 686.3  & \textbf{113439}  & 113500.3 & 29.3 & 870.9         \\
bio-yeast           & 1458   & 1948     & 26910     & 27091.5 & 85.45  & 380.23 & \textbf{26285}   & 26285    & 0        & 0.6           \\ \hline
ca-AstroPh          & 17903  & 196972   & n/a       & n/a     & n/a    & n/a    & \textbf{134376}  & 134418.9 & 32.27 & 966.93        \\
ca-citeseer         & 227320 & 814134   & n/a       & n/a     & n/a    & n/a    & \textbf{2928551} & 2940554.6  & 4349.55 & 990.2         \\
ca-coauthors-dblp   & 540486 & 15245729 & n/a       & n/a     & n/a    & n/a    & \textbf{2584027} & 2595357.9  & 3837.47 & 922.5         \\
ca-CondMat          & 21363  & 91286    & n/a       & n/a     & n/a    & n/a    & \textbf{207066}  & 207176.9 & 63.56 & 942.5         \\
\hline
\end{tabular}
\label{tab_8}
\end{table}

\begin{table}[htbp]
\centering
\renewcommand{\arraystretch}{1}
\caption{Experiment results of ACO-PP-LS and CC$^{2}$FS on the massive graph. {\small For the ith vertex $v_{i}$, $w(v_{i})$=($i$ mod 200)+1}}
\scriptsize\setlength{\tabcolsep}{4pt}
\begin{tabular}{|l|ll|llll|llll|}
\hline
\multicolumn{1}{|c|}{\multirow{2}{*}{INSTANCE}} & \multicolumn{1}{c}{\multirow{2}{*}{|V|}} & \multicolumn{1}{c|}{\multirow{2}{*}{|E|}} & \multicolumn{4}{c|}{ACO-PP-LS}                                                                          & \multicolumn{4}{c|}{CC$^{2}$FS}                                                                               \\
\multicolumn{1}{|c|}{}                          & \multicolumn{1}{c}{}                     & \multicolumn{1}{c|}{}                     & \multicolumn{1}{c}{MIN} & \multicolumn{1}{c}{AVG} & \multicolumn{1}{c}{SD} & \multicolumn{1}{c|}{Rtime} & \multicolumn{1}{c}{MIN} & \multicolumn{1}{c}{AVG} & \multicolumn{1}{c}{SD} & \multicolumn{1}{c|}{Rtime} \\ \hline

soc-BlogCatalog     & 88784  & 2093195  & n/a       & n/a     & n/a    & n/a    & \textbf{383083}  & 383122.5 & 47.7 & 968.91        \\
soc-brightkite      & 56739  & 212945   & n/a       & n/a     & n/a    & n/a    & \textbf{978395}  & 978472   & 106.93 & 893.68        \\
soc-buzznet         & 101163 & 2763066  & n/a       & n/a     & n/a    & n/a    & \textbf{7048}    & 7050.4   & 2.94 & 391.26        \\
soc-delicious       & 536108 & 1365961  & n/a       & n/a     & n/a    & n/a    & \textbf{4929139} & 4929250.4  & 68.94 & 634.44        \\
soc-digg            & 770799 & 5907132  & n/a       & n/a     & n/a    & n/a    & \textbf{5500630} & 5502484.2  & 894.5 & 970.33        \\
soc-douban          & 154908 & 327162   & n/a       & n/a     & n/a    & n/a    & \textbf{809539}  & 809674.8 & 61.52 & 838.91        \\
\hline

\end{tabular}
\label{tab_88}
\end{table}

\begin{table}[htbp]
\centering

\caption{Experiment results of CC$^{2}$FS, CC$^{2}$+S1, CC$^{2}$+S2, CC$^{2}$+S3, and CC$^{2}$+S4.}
\scriptsize\setlength{\tabcolsep}{3pt}
\begin{tabular}{|l|ll|ll|ll|ll|ll|}
\hline
Instance         & \multicolumn{2}{c|}{CC$^{2}$FS}            & \multicolumn{2}{c}{CC$^{2}$+S1}     & \multicolumn{2}{c|}{CC$^{2}$+S2}    & \multicolumn{2}{c|}{CC$^{2}$+S3}   & \multicolumn{2}{c|}{CC$^{2}$+S4}  \\
BHOSLIB          & MIN              & AVG           & MIN     & AVG            & MIN     & AVG             & MIN     & AVG            & MIN     & AVG          \\ \hline
frb45-21-2&\textbf{259}&259.3&268&269.2&272&279.9&277&283.3&320&321.3\\
frb50-23-2&\textbf{277}&277&278&281.6&281&282.5&325&325.5&341&341.6\\
frb53-24-2&\textbf{298}&300.3&314&314&323&328.6&328&333.6&344&349.6\\
frb56-25-2&\textbf{319}&319&328&328&340&340&341&341&360&360\\
frb59-26-2&\textbf{383}&388.8&410&416.8&426&428.4&444&447.2&472&474.4\\\hline
DIMACS&&&&&&&&&&\\
\hline
\end{tabular}
\label{tab_9}
\end{table}

\subsection{Results on BHOSLIB Benchmarks}
%what is meaning of the first sentence here?
Tables~\ref{tab_4} and \ref{tab_5} compare CC$^{2}$FS with ACO-PP-LS on the two weighted BHOSLIB benchmarks, one of which adopts the weighting function $w(v_{i})$=($i$ mod 200)+1 (Table~\ref{tab_4}), while the other adopts the weighting function $w(v_{i})$=1 (Tables~\ref{tab_5}). We compare the number of optimal solutions found by the two algorithms. Once again, CC$^{2}$FS dramatically outperforms its competitors ACO-PP-LS. All instances of this benchmark could be solved by our algorithm more quickly than ACO-PP-LS. More importantly, our algorithm could find better solution values.

\subsection{Results on DIMACS Benchmarks}
The experimental results on the two weighted DIMACS benchmarks are presented in Tables~\ref{tab_6} and \ref{tab_7}. It is encouraging to see the performance of CC$^{2}$FS remains surprisingly good on these instances, where ACO-PP-LS shows very poor performance. Despite the performance advantage of CC$^{2}$FS, there are some exceptions in the weighting function of $w(v_{i})$=($i$ mod 200)+1, like gen200\_p0.9\_44 and C500.9. For the weighting function $w(v_{i})$=1, in the Table~\ref{tab_7}, our algorithms has a significant improvement than ACO-PP-LS in terms of real run time and best solution value. Given the good performance of CC$^{2}$FS on the DIMACS benchmark with large vertices, we are confident it could be able to solve larger graph instances.

\subsection{Results on Massive Graph Benchmark}
For some instances, ACO-PP-LS fails to find a dominating set within time limit, then we use ``n/a'' to mark it.
From the results in Table~\ref{tab_8} and Table~\ref{tab_88}, we observe that there are 58 graphs for which ACO-PP-LS fails to provide a dominating set, which indicates the effectiveness of our algorithm. In the rest instances (16 ones), CC$^{2}$FS finds better dominating sets than ACO-PP-LS. Moreover, CC$^{2}$FS could find the good dominating sets of all given massive graphs (74 ones).

\SetAlFnt{\small}

\begin{algorithm}[t]

\caption{CC$^{2}$FS+BREAK ($G$, {\it cutoff})}
\KwIn{a weighted graph $G=(V,E,W)$, the {\it cutoff} time }

\KwOut{dominating set of $G$ }

initialize $ConfChange$ and $forbid\_list$\;
initialize the $freq$ and $score_{f}$ of of vertices\;
$S := InitGreedyConstruction()$\;
$S^{*} := S$\;

%$step:=0$;
%$C^{*} :=  \emptyset $
%$C:=ConstructVC()$\;
%$gain(v):=0$ for each vertex $v\notin C$\;

    \While{elapsed time $<$ {\it cutoff} }
    {
      \If{there are no uncovered vertices} {
         $S^{*} := S$\;
         $v :=$ a vertex in $S$ with the highest value $score_{f}(v)$, breaking ties in the oldest one\;
         $S := S \setminus \{v\}$\;
         update $ConfChange$ according to CC$^{2}$-RULE2\;
         continue\;
      }
       $v :=$ a vertex in $S$ with the highest value $score_{f}(v)$ and $v \notin forbid\_list$, breaking ties in the oldest one\;
      $S := S \setminus \{v\}$\;
      update $ConfChange$ according to CC$^{2}$-RULE2\;
      $forbid\_list := \emptyset $\;
      \While{ there are no uncovered vertices}
      {
         $v :=$ a vertex in $CCV^{2}$ with the highest value $score_{f}(v)$ and $ConfChange$[$v$]$\neq$false, breaking ties in the oldest one\;
         \lIf{$ w(S)+w(v)\geq w(S^{*})$} {break}
         $S := S \cup \{v\}$\;
         update $ConfChange$ according to CC$^{2}$-RULE3\;
         $forbid\_list := forbid\_list \cup \{v\} $\;
         $freq[v] := freq[v]+1$, for $v \notin N[S]$\;

      }

    }

\Return $ S^{*}$\;

\end{algorithm}

\subsection{Comparison Frequency based Scoring Function with Previous Scoring Functions}
To study the effectiveness of frequency based scoring function, we use four scoring functions \cite{ref17} instead of frequency based scoring function and design four alternative algorithms:
CC$^{2}$+S1 which uses $score_{1}$ to select vertices,
CC$^{2}$+S2 based on $score_{2}$,
CC$^{2}$+S3 with $score_{3}$, as well as
CC$^{2}$+S4 by $score_{4}$.

In Table~\ref{tab_9}, we can easily observe that CC$^{2}$FS consistently obtains better solutions than other competitors for all selected instances. It demonstrates that our proposed frequency based scoring function is suitable for this problem and helps CC$^{2}$FS find better solutions than previous scoring functions.

\subsection{Comparison with the Original Configuration Checking and the Breaking Strategy}
To study the effectiveness of the CC$^2$ strategy, we compare CC$^2$FS with its alternative version named CCFS, which uses one-level neighborhood based configuration of vertices.
%Also, in CC$^2$FS, we add one vertex into the candidate solution until this candidate solution is a dominating set.

We also compare CC$^2$FS with another version CC$^{2}$FS+BREAK, which differs from CC$^2$FS on the stopping criterion of adding vertices in each step: it stops the adding process when either the candidate solution becomes a dominating set, or the cost of the candidate solution is larger than that of the best found solution. The pseudo code of CC$^{2}$FS+BREAK is introduced in Algorithm 3.
Although CC$^{2}$FS is similar to CC$^{2}$FS+BREAK, there are two differences between CC$^{2}$FS and CC$^{2}$FS+BREAK.

\begin{itemize}
  \item (1) When each iteration begins, both algorithms check whether there are no uncovered vertices (line 6). If this is the case, CC$^{2}$FS+BREAK finds a better solution and updates $S^{*}$ by the candidate solution $S$, while CC$^{2}$FS cannot ensure that $S$ is better than $S^{*}$ and this has to be checked.
  \item (2) In the inner loop (lines 12-16 in CC$^{2}$FS, lines 16-22 in CC$^{2}$FS+BREAK), CC$^{2}$FS+BREAK turns to the next iteration when there are no uncovered vertices (line 16) or $ w(S)+w(v)\geq w(S^{*})$ (line 18), while our algorithm CC$^{2}$FS can only continue to the next iteration under meeting the first condition.
\end{itemize}

 %If our algorithm judge whether the sum of weight of vertices in the candidate solution is larger than the total weight of local optimal found by the previous search of our algorithm, then our algorithm will break and turn to the next iteration. Based this break strategy, we propose another algorithm called CC$^{2}$FS+BREAK.

The results are summarized in Table~\ref{tab_10}, which shows that the CC$^{2}$FS finds better solutions than CCFS on all benchmarks. This indicates that the new configuration checking CC$^{2}$ plays a key role in the CC$^{2}$FS algorithm. Compared with CC$^{2}$FS+BREAK, CC$^{2}$FS could get better or the same quality solutions.
This indicates that the break strategy is not suitable for this problem.
Furthermore, the visualized comparisons of CC$^{2}$FS, CC$^{2}$FS+BREAK, and CCFS can be seen by box-plot in Figure 1, which shows the distribution of the dominating set values of massive graphs.

\begin{table}[htbp]
\centering
\caption{Experiment results of CC$^{2}$FS, CC$^{2}$FS+BREAK, and CCFS. For CC$^{2}$FS and CCFS, the ADDset column reports the number of vertices which are allowed to be added into the candidate solution.}

\scriptsize\setlength{\tabcolsep}{6pt}
\begin{tabular}{|l|lll|ll|lll|}
\hline
\multicolumn{1}{|c|}{\textbf{Instance}} & \multicolumn{3}{c|}{\textbf{ CC$^{2}$FS}}                                                                           & \multicolumn{2}{c|}{\textbf{CC$^{2}$FS+BREAK}}                           & \multicolumn{3}{c|}{\textbf{CCFS}}                                                                            \\ \cline{1-1}
\multicolumn{1}{|c|}{\textbf{T1}}       & \multicolumn{1}{c}{\textbf{MEAN}} & \multicolumn{1}{c}{\textbf{Rtime}} & \multicolumn{1}{c|}{\textbf{ADDset}} & \multicolumn{1}{c}{\textbf{MEAN}} & \multicolumn{1}{c|}{\textbf{Rtime}} & \multicolumn{1}{c}{\textbf{MEAN}} & \multicolumn{1}{c}{\textbf{Rtime}} & \multicolumn{1}{c|}{\textbf{ADDset}} \\ \hline
v300e300                                & \textbf{3178.6}                   & 1.47                               & \textbf{177.6}                       & 3183                              & 0.57                                & 3178.7                            & 2.43                               & 176.2                                \\
v500e500                                & \textbf{5305.7}                   & 2.63                               & \textbf{296.8}                       & 5327.2                            & 0.92                                & 5307.3                            & 4.73                               & 295.4                                \\
v800e1000                               & \textbf{7663.4}                   & 10.58                              & \textbf{504.1}                       & 7714.1                            & 2.17                                & 7673.7                            & 8.12                               & 502.5                                \\
v800e2000                               & \textbf{4982.1}                   & 8.83                               & \textbf{600.4}                       & 5005.3                            & 3.59                                & 4989.9                            & 11.67                              & 598.2                                \\
v1000e1000                              & \textbf{10585}                  & 12.18                              & \textbf{590.9}                       & 10689.6                           & 5.15                                & 10588.4                           & 16.68                              & 589.5                                \\ \hline
T2                                      & MEAN                              & Rtime                              & ADDset                               & MEAN                              & Rtime                               & MEAN                              & Rtime                              & ADDset                               \\ \hline
v200e2000                               & \textbf{884.6}                    & 0.09                               & \textbf{175.5}                       & 887.1                             & 0.17                                & 884.6                             & 0.01                               & 171.5                                \\
v500e500                                & \textbf{623.6}                    & 0.29                               & \textbf{285.2}                       & 626.5                             & 1.03                                & 623.9                             & 0.11                               & 283.8                                \\
v800e1000                               & \textbf{1104.3}                   & 2.36                               & \textbf{475.0}                       & 1110.9                            & 0.99                                & 1106.1                            & 2.5                                & 471.8                                \\
v800e2000                               & \textbf{1632.3}                   & 3.59                               & \textbf{549.5}                       & 1634.9                            & 0.84                                & 1642.2                            & 3.24                               & 539.6                                \\
v1000e1000                              & \textbf{1237.7}                   & 0.86                               & \textbf{569.6}                       & 1252                              & 2.72                                & 1238.3                            & 5.27                               & 567.9                                \\ \hline
BHOSLIB                                 & MIN                               & Rtime                              & ADDset                               & MIN                               & Rtime                               & MIN                               & Rtime                              & ADDset                               \\ \hline
frb45-21-2                              & \textbf{259}                      & 112.14                             & \textbf{914}                         & 260                               & 1.51                                & \textbf{259}                      & \textbf{3.94}                      & 912                                  \\
frb50-23-2                              & \textbf{277}                      & \textbf{0.02}                      & \textbf{1110}                        & \textbf{277}                      & 0.03                                & \textbf{277}                      & 0.04                               & 1108                                 \\
frb53-24-2                              & \textbf{298}                      & 0.33                               & \textbf{1228}                        & \textbf{298}                      & \textbf{0.04}                       & \textbf{298}                      & \textbf{0.16}                      & 1227                                 \\
frb56-25-2                              & \textbf{319}                      & \textbf{0.64}                      & \textbf{1360}                        & 323                               & 3.22                                & \textbf{319}                      & 1.36                               & 1358                                 \\
frb59-26-2                              & \textbf{383}                      & 146.19                             & \textbf{1498}                        & 394                               & 27.01                               & \textbf{383}                      & \textbf{35.92}                     & 1485                                 \\ \hline
DIMACS                                  & MIN                               & Rtime                              & ADDset                               & MIN                               & Rtime                               & MIN                               & Rtime                              & ADDset                               \\ \hline
gen200\_p0.9\_44                        & \textbf{470}                               & 2.99                               & \textbf{179}                         & 472                               & \textless0.01                       & \textbf{470}                      & 0.01                               & 174                                  \\
gen200\_p0.9\_55                        & \textbf{433}                      & \textbf{0.03}                      & \textbf{179}                         & 434                               & \textless0.01                       & \textbf{443}                      & 0.04                               & 173                                  \\
hamming8-4                              & \textbf{71}                       & 7.70                               & \textbf{243}                         & 74                                & 0.66                                & 73                                & 56.39                              & 242                                  \\
keller4                                 & \textbf{220}                      & 1.26                               & \textbf{162}                         & \textbf{220}                      & 0.07                                & \textbf{220}                      & \textbf{0.01}                      & \textbf{162}                         \\
p\_hat700-1.clq                         & \textbf{67}                       & 0.02                               & \textbf{682}                         & \textbf{67}                       & 0.02                                & \textbf{67}                       & 0.02                               & 680                                  \\ \hline
massive-graph                           & MIN                               & Rtime                              & ADDset                               & MIN                               & Rtime                               & MIN                               & Rtime                              & ADDset                               \\ \hline
bio-delma                               & \textbf{113439}                   & 870.9                              & \textbf{5623}                        & 113966                            & 749.13                              & 113560                            & 447.4                              & 5468                                 \\
ca-dblp-2010                            & \textbf{3471926}                  & 150.6                              & \textbf{188608}                      & 3471932                           & 87.7                                & 3471927                           & 135.3                              & 187496                               \\
ia-wiki-Talk                            & \textbf{972775}                   & 982.55                             & \textbf{78097}                       & 979393                            & 20.92                               & 972873                            & 987.13                             & 77159                                \\
inf-power                               & \textbf{121049}                   & 597.61                             & \textbf{3108}                        & 122229                            & 74.47                               & 121659                            & 59.16                              & 3070                                 \\
rec-amazon                         & \textbf{2092250}                  & 999.86                             & \textbf{56815}                       & 2140426                           & 26.31                               & 2092515                          & 993.11                             & 56539                                \\
sc-ldoor                                & \textbf{5459575}                  & \textbf{617.64}                             & \textbf{879199}                      &  \textbf{5459575}                           & 726.79                              & 5459782                           & 793.57                             & 878119                               \\
Soc-digg                                & \textbf{5500630}                  & 970.33                             & \textbf{688382}                      & 5500701                          & 920.32                              & 5533949                         & 980.23                             & 681599                               \\
socfb-Texas84                          & \textbf{118295}                   & 897.64                             & \textbf{33242}                       & 118373                            & 979.53                              & 118767                            & 970.12                             & 32790                                \\
Tech-RL-caida                           & \textbf{3151996}                  & 946.29                             & \textbf{141841}                      & 3185071                           & 180.93                              & 3164081                           & 992.34                             & 139402                               \\
web-arabic-2005                         & \textbf{1580428}                  & 992.23                             & \textbf{144628}                      & 1593262                           & 46.11                               & 1580762                          & 940.23                             & 143899                               \\ \hline
\end{tabular}
\label{tab_10}
\end{table}

\begin{figure}[!h]
\centering
%\vspace{1.8cm}
\includegraphics[scale=0.65]{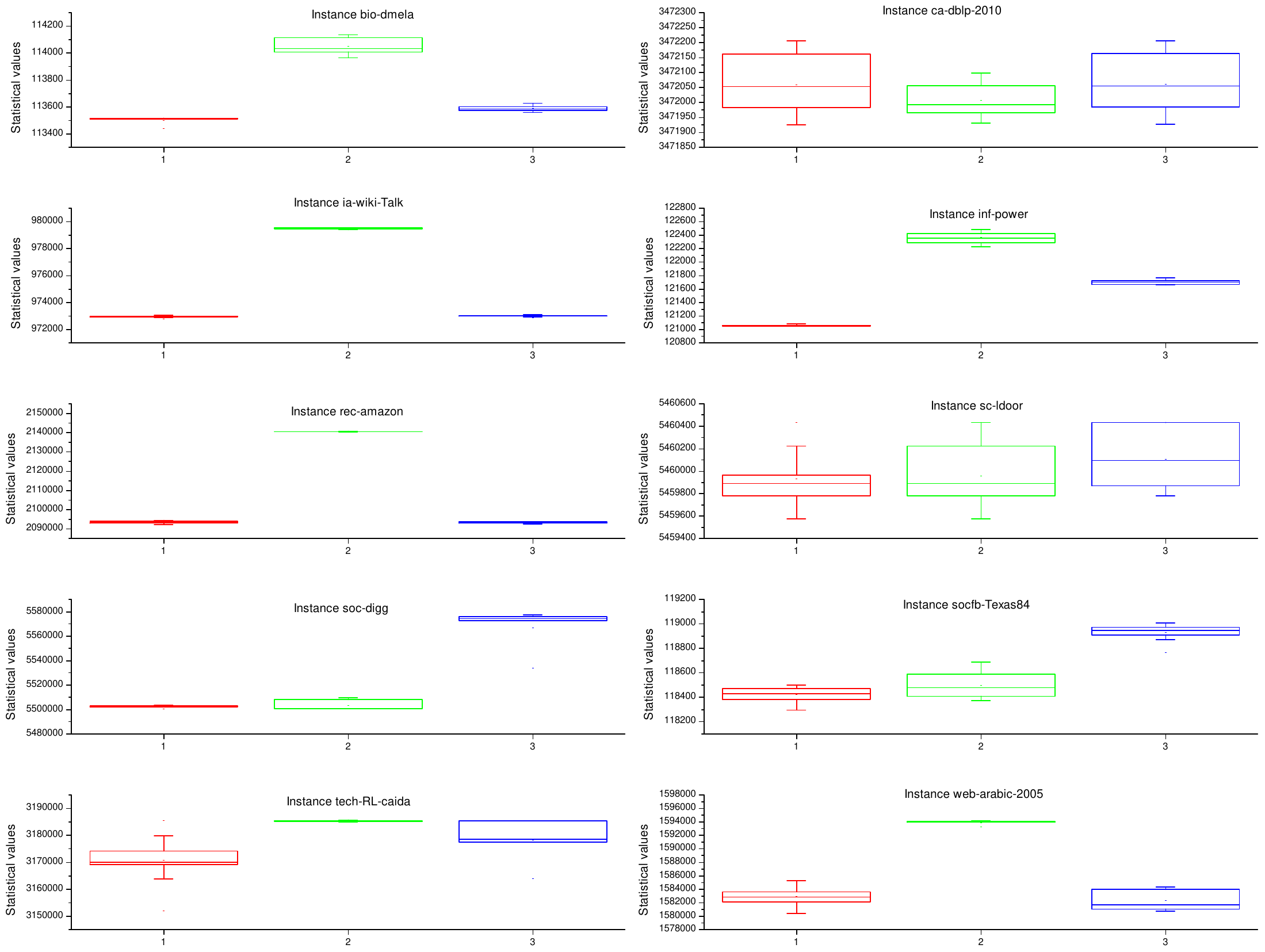}
\caption{The dominate set values of massive graphs obtained by 1: CC$^{2}$FS; 2: CC$^{2}$FS+BREAK; 3: CCFS. \small{Box plots are applied to display the distribution of these values.}}
\label{fig:example1}
\end{figure}

\section{Summary and Future Work}
This paper presented a local search algorithm called CC$^{2}$FS for solving the minimum weight dominating set (MWDS) problem. We proposed a new configuration checking strategy namely CC$^{2}$ based on the two-level neighborhood of vertices to remember the relevant information of removed and added vertices and prevent visiting the recent paths. Moreover, we introduced a new frequency based scoring function for solving MWDS. The experimental results showed that CC$^{2}$FS performs essentially better than state of the art algorithms on almost all instances in terms of solution quality and run time.

As for future work, we consider to further improve the CC$^{2}$FS algorithm by integrating some other ideas like strong configuration checking \cite{ref87}. Also we would like to test our algorithms on other instances including larger graphs. Envisioned research directions about the proposed strategies include applying the new score functions to other local search algorithms, and trying to find some other important properties and scoring functions of local search algorithms.

%\section*{Acknowledgement}

\newpage

\acks{The authors of this paper wish to extend their sincere gratitude to all the anonymous reviewers for their efforts.
Shaowei Cai is also supported by Youth Innovation Promotion Association, Chinese Academy of Sciences.
For any theoretical and experimental problem arising from this paper, please correspondence to Professor Minghao Yin. 
This work was supported in part by NSFC (under Grant Nos. 61370156, 61503074, 61502464, 61402070, 61403077, and 61403076), China National 973 program 2014CB340301 and the Program for New Century Excellent Talents in University (NCET-13-0724). 
}

\vskip 0.2in
\bibliography{sample}
\bibliographystyle{theapa}

\end{document}